\documentclass[runningheads]{llncs}

\usepackage{eccv}

\usepackage{eccvabbrv}

\usepackage{graphicx}
\usepackage{booktabs}
\usepackage{pgfplots}
\usepackage[table]{xcolor}
\usepackage{pgfplots}
\pgfplotsset{compat=1.18}
\usepackage{multirow}
\usepackage[misc]{ifsym}
\usepackage[accsupp]{axessibility}  
\AtBeginDocument{
  \setlength\abovedisplayskip{7pt plus 2pt minus 3pt}
  \setlength\belowdisplayskip{7pt plus 2pt minus 3pt}
  
  \setlength\abovedisplayshortskip{5pt plus 2pt minus 2pt}
  \setlength\belowdisplayshortskip{5pt plus 2pt minus 2pt}
}

\definecolor{mygray}{gray}{.92}
\makeatletter
\def\blfootnote{\xdef\@thefnmark{}\@footnotetext}
\makeatother

\usepackage{hyperref}

\usepackage{orcidlink}

\begin{document}

\title{LightSTAR: Efficient Visual Document \\ Retrieval via Lightweight Selection with \\Vision-Adaptive Refinement} 

\titlerunning{LightSTAR for Efficient Visual Document Retrieval} 

\author{Tongkun Guan\inst{1}\textsuperscript{$*$}\orcidlink{0000-0003-3346-8315} \and
Haocheng Wang\inst{1}\textsuperscript{$*$}\orcidlink{0009-0008-8449-1834} \and
Wei Shen\inst{1(\textrm{\Letter})} \and
Xiaokang Yang\inst{1}
}

\authorrunning{T.~Guan et al.}

\institute{MoE Key Lab of Artificial Intelligence, AI Institute, School of Computer Science, \\ Shanghai Jiao Tong University \email{\{gtk0615,wanghaocheng2023\}@sjtu.edu.cn} \\
}



\maketitle
\blfootnote{\noindent$^{*}$Equal contribution. \textsuperscript{\Letter}Corresponding author.}

\begin{abstract}
 Visual document retrieval requires rapidly locating relevant pages from large multi-modal corpora in response to user queries. While recent methods powered by Multi-modal Large Language Models (MLLMs) show competitive accuracy, they suffer from prohibitive computational costs by applying intensive MLLM encoding to every single page. Meanwhile, we observe that user queries are typically keyword-anchored, containing semantically rich words that are expected to appear directly in the visible text of relevant pages, offering an efficient cue for quickly narrowing down candidate pages. Building on this insight, we propose LightSTAR, an efficient framework that decomposes visual document retrieval into: 1) LLM-free Visual Selection, which utilizes content-grounded query encoding to focus on informative words and employs LLM-free visual embeddings to produce a high-recall candidate set; and 2) Vision-adaptive Semantic Refinement, which further performs fine-grained semantic matching exclusively on these top candidates via adaptive region-wise feature fusion to effectively combine textual and layout cues, optimized through a hardness-aware contrastive objective. Experimental results demonstrate that LightSTAR achieves state-of-the-art retrieval accuracy while reducing end-to-end latency by several-fold, offering a highly practical solution to the accuracy-efficiency trade-off in visual document retrieval. Code is available at \url{https://github.com/bokufa/LightSTAR}.
\end{abstract}

\section{Introduction}
\label{sec:intro}
The explosion of digital documents (from lengthy legal contracts to technical reports filled with diagrams and tables) is pushing retrieval systems beyond plain text. Modern applications such as retrieval-augmented generation (RAG)~\cite{lewis2021retrievalaugmentedgenerationknowledgeintensivenlp}, enterprise search, and scientific assistance require processing hundreds or thousands of pages where crucial information is encoded not only in text but also in visual structure: layouts, figures, tables, and typography. This shift calls for retrieval systems capable of rapidly locating relevant pages from large multi-modal corpora. Such systems must not only understand complex visual semantics, but also operate efficiently at scale.

Traditional text-based retrieval systems struggle when crucial information is embedded in rich visual and structural elements, as they rely on imperfect text extraction and often lose critical layout and visual cues essential for understanding document semantics~\cite{zhang2025ocrhindersragevaluating}. This gap has motivated the development of visual document retrieval~\cite{faysse2025colpaliefficientdocumentretrieval}, which operates directly on document images to preserve native layout and multi-modal cues. 
Recent breakthroughs in multi-modal large language models (MLLMs)~\cite{li2022blipbootstrappinglanguageimagepretraining,liu2023visualinstructiontuning,alayrac2022flamingovisuallanguagemodel} have enabled sophisticated approaches to this challenge. Methods such as ColPali~\cite{faysse2025colpaliefficientdocumentretrieval} and VisRAG~\cite{yu2025visragvisionbasedretrievalaugmentedgeneration} leverage powerful MLLMs to embed document images and text queries into unified semantic spaces, employing late interaction mechanisms~\cite{khattab2020colbertefficienteffectivepassage, li2022blipbootstrappinglanguageimagepretraining, radford2021learningtransferablevisualmodels} to capture fine-grained cross-modal correspondences. While these MLLM-based approaches achieve competitive retrieval accuracy on challenging benchmarks~\cite{faysse2025colpaliefficientdocumentretrieval, muennighoff2023mtebmassivetextembedding, günther2025jinaembeddingsv4universalembeddingsmultimodal}, a major obstacle remains: they require MLLM-level computation for every page during indexing and retrieval. For long documents and frequently updated corpora, encoding all pages with a full MLLM is prohibitively expensive in latency, throughput, and cost~\cite{kaplan2020scalinglawsneurallanguage, brown2020languagemodelsfewshotlearners, dao2022flashattentionfastmemoryefficientexact}. This inefficiency stems from the fact that, for any given query, only a tiny fraction of pages are potentially relevant, yet existing systems apply the same heavy computation to the vast majority of obviously irrelevant pages. This calls for a selective computation in which inexpensive signals are used to prune large corpora before invoking costly multi-modal reasoning.

This paper is motivated by a key observation from real-world document retrieval scenarios: user queries are typically keyword-anchored~\cite{thakur2021beirheterogenousbenchmarkzeroshot, zhang2025ocrhindersragevaluating,10.1145/1571941.1571989}. They contain a compact set of semantically rich content words (specific entities, domain terminology, distinctive actions, or characteristic attributes) that are expected to appear on relevant pages, while most other pages lack such cues and can be filtered using much cheaper visual-text signals. 
 This suggests that expensive MLLM computation can be reserved for refining a compact candidate set identified through lightweight filtering. Based on this principle, we propose \textbf{LightSTAR} (\textbf{Light}weight \textbf{S}elec\textbf{T}ion with vision-\textbf{A}daptive \textbf{R}efinement for efficient visual document retrieval), a novel framework that strategically allocates computation to where it matters. LightSTAR decomposes visual document retrieval into: \textbf{1) LLM-free Visual Selection.} 
We develop a lightweight, text-aware visual encoder, where query can directly interact with document images with a shared embedding space, to perform fast corpus-wide filtering and produce a small candidate set without introducing any large language model (LLM). To enhance retrieval effectiveness, we propose content-grounded query embedding, which applies linguistic analysis to filter semantically informative tokens while removing function words that are visually uninformative and ubiquitously distributed across documents. The filtered query tokens are then matched against document patch embeddings via a novel scale-adaptive late interaction scheme, enabling efficient, high-recall pruning at low cost. \textbf{2) Vision-adaptive Semantic Refinement.}
On the filtered candidates, we perform MLLM-based refinement for fine-grained semantic matching. We reuse the text-aware visual encoder to obtain semantically aligned representations. However, text-aware features alone may overlook complementary background and layout cues. To address this, we introduce an adaptive region-wise feature fusion strategy that preserves text-aware features on foreground/text regions while injecting layout-aware features on background/layout regions.
Moreover, since candidates contain similar pages, we adopt a hardness-aware contrastive objective that emphasizes confusable negatives, encouraging more discriminative fine-grained representations.

Finally, our main contributions can be summarized as twofold:
\begin{itemize}
    \item We introduce LightSTAR, which strategically decouples visual document retrieval into fast corpus-wide filtering via a lightweight LLM-free Visual Selection module and fine-grained candidate matching via Vision-adaptive Semantic Refinement.
    \item Experimental results demonstrate that LightSTAR achieves state-of-the-art retrieval quality while significantly improving latency efficiency compared to existing MLLM-based methods, making it practical for real-world deployments with large-scale document collections. 
\end{itemize}
\section{Related Work}
\label{sec:related}

\noindent \textbf{Vision-language Models.} Vision-language models have achieved remarkable progress in document understanding tasks~\cite{wen2026efficient, tschannen2025siglip,Guan_2023_CVPR,kirillov2023segment,simeoni2025dinov3,fu-etal-2025-multimodal,wang2025marten,guan2026codepercept,guan2025tokenleveltextimagefoundation,guan2024bridging,guan2023self_,guan2025posformer,guan2023self,guan2022industrial,guan2025ccdplus}. Early works such as LayoutLM~\cite{Xu_2020} and LayoutLMv2~\cite{xu2022layoutlmv2multimodalpretrainingvisuallyrich} incorporate layout information by combining visual features with text embeddings extracted from OCR. More recent approaches adopt an OCR-free paradigm: Donut~\cite{kim2022ocrfreedocumentunderstandingtransformer} employs an encoder-decoder architecture to directly generate text from document images, while Pix2Struct~\cite{lee2023pix2structscreenshotparsingpretraining} pretrains on page screenshots for structured understanding. TokenFD~\cite{guan2025tokenleveltextimagefoundation} further advances OCR-free document understanding by introducing a token-level approach for fine-grained document feature extraction.
The emergence of powerful generalist MLLMs—such as LLaVA~\cite{liu2023visualinstructiontuning}, PaliGemma~\cite{beyer2024paligemmaversatile3bvlm}, Qwen-VL~\cite{bai2023qwenvlversatilevisionlanguagemodel}, and InternVL~\cite{chen2024internvlscalingvisionfoundation}—has further advanced document understanding capabilities. These models leverage large-scale pretraining to achieve strong vision-language alignment, enabling them to understand complex document layouts, extract textual content, and reason over visual elements.


\noindent \textbf{Visual Document Retrieval.} Visual document retrieval aims to retrieve relevant documents given text queries, where documents are represented as images containing mixed modalities—text, tables, figures, and complex layouts interleaved on a single page. Early approaches typically extract text via OCR engines and then apply text-based retrieval methods~\cite{robertson1994okapi,10.1108/eb026526}. However, such pipelines suffer from OCR errors~\cite{zhang2025ocrhindersragevaluating} and fail to capture visual layout information that is crucial for document understanding. Recent advances leverage multi-modal large language models to directly encode document images without relying on external OCR. ColPali~\cite{faysse2025colpaliefficientdocumentretrieval} employs PaliGemma to generate multi-vector representations for document images and adopts late interaction~\cite{khattab2020colbertefficienteffectivepassage} for query-document matching, achieving strong retrieval performance. Follow-up works extend this paradigm with different MLLM backbones: ColQwen, a variant of ColPali, builds upon Qwen-VL~\cite{bai2023qwenvlversatilevisionlanguagemodel} for enhanced multilingual support. VisRAG~\cite{yu2025visragvisionbasedretrievalaugmentedgeneration} investigates single-vector representations as an alternative to multi-vector approaches. Despite their effectiveness, these methods uniformly require LLM-based encoding for the entire corpus during indexing. Our work addresses this limitation via the combination of a lightweight LLM-free Visual Selection module (efficiently filters candidates) and a MLLM-based Vision-Adaptive Refinement module (distinguishes the subtle semantic differences of the candidates), achieving both scalability and accuracy.


\section{Methodology}
\label{sec:method}
\noindent \textbf{Problem Formulation.} 
Visual document retrieval has emerged as a critical capability for processing real-world multi-modal documents (\eg, technical reports, legal contracts, and scientific manuscripts) that often span hundreds or thousands of visually rich pages. For instance, in RAG pipelines, efficient first-stage evidence selection over large page collections is a prerequisite for any downstream reasoning, making the ability to rapidly and accurately identify relevant pages from large-scale document corpora directly determines the effectiveness and scalability of the entire system.

Specifically, we formalize this visual document retrieval task as a ranking problem over document images. Given a natural language query $q$ and a collection of document pages $\mathcal{D} = \{d_1, d_2, \ldots, d_N\}$, where each $d_i$ represents a document image containing rich visual and textual information, the objective of visual document retrieval is to rank all pages $\mathcal{D}$ by their relevance to the query:
\begin{equation}
\label{eq:ranking}
    [d_{\pi_1}, d_{\pi_2}, \ldots, d_{\pi_N}] = \operatorname*{argsort}_{d_i \in \mathcal{D}} \mathcal{S}(q, d_i),
\end{equation}
where $\mathcal{S}: (q, d_i) \rightarrow \mathbb{R}$ is a similarity scoring function that quantifies the semantic correspondence between query $q$ and document page $d_i$. $\pi$ is a permutation such that $\mathcal{S}(q, d_{\pi_1}) \geq \mathcal{S}(q, d_{\pi_2}) \geq \cdots \geq \mathcal{S}(q, d_{\pi_N})$.

To address this problem, existing approaches predominantly rely on encoding each document page with powerful MLLMs~\cite{faysse2025colpaliefficientdocumentretrieval, yu2025visragvisionbasedretrievalaugmentedgeneration}, leveraging their sophisticated capacity for fine-grained visual understanding to produce rich semantic representations. While this strategy effectively yields superior retrieval quality, it introduces a critical computational bottleneck: applying large-scale MLLMs to every page in a long document results in substantial computational overhead and prohibitive latency, limiting its practicality for large-scale or time-sensitive applications. This motivates our focus on retrieval methods that efficiently narrow the search space over long multi-modal documents without sacrificing the semantic fidelity required for accurate retrieval.

\begin{figure}[t]
    \centering
    \includegraphics[width=1\linewidth]{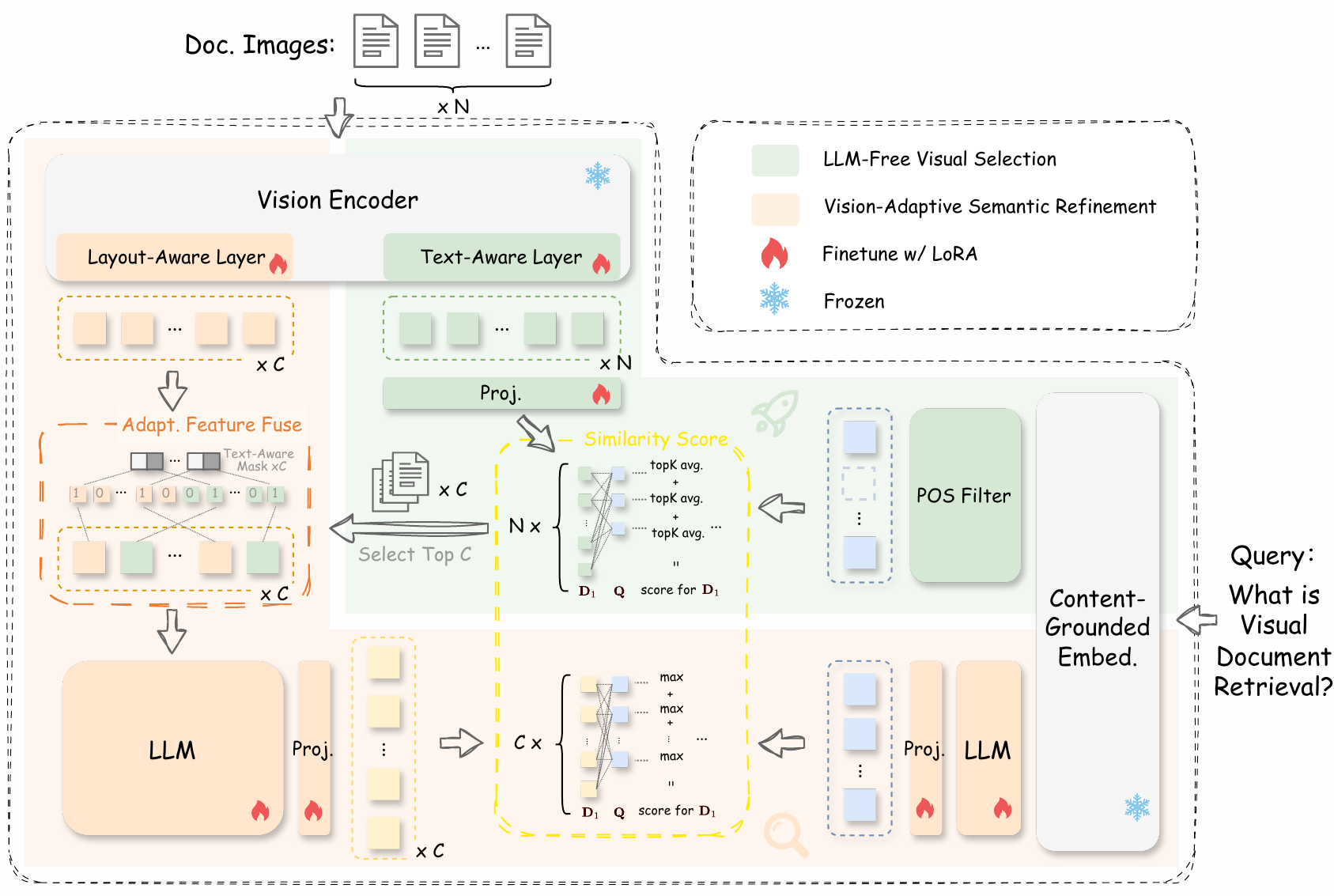}
    \vspace{-1.5em}
    \caption{Overview of the proposed method LightSTAR. Our method decomposes retrieval into LLM-Free Visual Selection and Vision-Adaptive Semantic Refinement, achieving state-of-the-art accuracy with lowest latency cost.}
    \label{fig:framework}
    \vspace{-1.5em}
\end{figure}

\subsection{Overview} 
\label{sec:overview}
As formulated above, the core challenge lies in balancing retrieval accuracy against computational efficiency when processing long multi-modal documents. To address this, we begin with a key empirical observation: in practical scenarios, user queries typically correspond to only a small subset of pages within a document collection. Because queries are often keyword-anchored and contain distinctive textual cues such as domain-specific terms expected on relevant pages, pages lacking these fundamental cues can be efficiently filtered using accurate visual signals, obviating the need for expensive semantic modeling. This observation suggests that by first identifying a small set of likely relevant pages using inexpensive visual-text signals, expensive full semantic modeling can be selectively applied only where it is truly needed.

Building on this insight, we propose LightSTAR, an effective framework that strategically decouples retrieval into: 

\noindent \textbf{LLM-free Visual Selection.} Without introducing any LLM, we develop a lightweight, text-aware visual encoder to rapidly filter the document collection and identify a promising candidate set $\mathcal{D}_c \subset \mathcal{D}$ where $|\mathcal{D}_c| \ll |\mathcal{D}|$, achieving computational complexity of $\mathcal{O}(N \cdot C_{\text{VE}})$ with $C_{\text{VE}} \ll C_{\text{MLLM}}$. $C_{\text{VE}}$ and $C_{\text{MLLM}}$ denote the complexity of the visual encoder and MLLM, respectively;

\noindent \textbf{Vision-adaptive Semantic Refinement.} The filtered candidate set $\mathcal{D}_c$ is further processed by a powerful MLLM to perform fine-grained semantic matching, reducing complexity to $\mathcal{O}(|\mathcal{D}_c| \cdot C_{\text{MLLM}})$ where $|\mathcal{D}_c| \ll N$. Additionally, it shares the same visual encoder as the first-stage LLM-free Visual Selection module, eliminating redundant parameters and lowering the memory footprint, which further enhances computational efficiency.

This design achieves overall complexity of $\mathcal{O}(N \cdot C_{\text{VE}} + |\mathcal{D}_c| \cdot C_{\text{MLLM}})$, yielding substantial speedups when $|\mathcal{D}_c| / N$ is small, while retaining most of the semantic expressiveness compared to full MLLM-based retrieval with complexity $\mathcal{O}(N \cdot C_{\text{MLLM}})$.
This architecture provides a practical approach to managing the efficiency–quality trade-off in large-scale retrieval within long multi-modal documents. By strategically allocating computation where it matters most, LightSTAR fully leverages the efficiency of lightweight encoders and the accuracy of MLLMs, making it practical for large-scale deployment. Figure~\ref{fig:framework} illustrates the complete pipeline, where the two modules synergistically combine efficiency and accuracy to enable practical large-scale visual document retrieval.

\subsection{LLM-free Visual Selection}
\label{sec:stage1}
To reduce computational overhead in large-scale document image retrieval, this module performs rapid candidate selection over the entire document collection without introducing any large language model. Instead, it relies exclusively on lightweight text-aware visual representations that capture visual-text signal for relevance estimation.

\noindent \textbf{Content-grounded Query Embedding.}
In real-world retrieval scenarios, user queries are typically formulated to explicitly specify the key information needed to locate target documents. Such queries naturally contain semantically rich content words expected to appear in the visual textual content of relevant pages~\cite{thakur2021beirheterogenousbenchmarkzeroshot, 10.1145/1571941.1571989,10.1561/1500000019, 10.1108/eb026526}. We also conduct an analysis on existing datasets which shows that the vast majority of queries contain visually observable content tokens in their ground-truth pages (Please see Appendix for details). However, natural language queries inevitably include function words (\eg, \textit{the}, \textit{of}, \textit{and}, \textit{to}, \textit{is}) that, while grammatically necessary, carry minimal semantic content and rarely contribute to visual document matching. 
Function words introduce noise in visual matching because they appear ubiquitously across documents regardless of relevance, creating spurious similarity signals. Treating function words equally with content words dilutes the discriminative power of query representations.

To obtain a content-grounded query representation, we apply part-of-speech (POS) tagging~\cite{loper2002nltknaturallanguagetoolkit} to extract semantically informative tokens while filtering out function words. Specifically, we retain only semantically informative tokens while discarding function words unlikely to contribute to visual relevance:
\begin{align}
\label{eq:query_filter}
    q' = \{\, t_i \in q \mid 
    \mathrm{POS}(t_i) \notin \{\mathrm{CC}, \mathrm{IN}, \mathrm{DT}\}
    \ \land\ 
    t_i \notin \{\mathcal{B}, \mathcal{P}\}
    \},
\end{align}
where $\text{POS}(\cdot)$ denotes the POS tagger output. The set \(\{\mathrm{CC}, \mathrm{IN}, \mathrm{DT}\}\) contains the excluded POS categories (coordinating conjunctions, prepositions/subordinating conjunctions, and determiners). \(\mathcal{B}\) and \(\mathcal{P}\) denote predefined sets of \emph{be}-verbs and punctuation tokens, respectively.
The filtered query $q'$ is encoded via embedding lookup as $\mathbf{Q}$, each retained word $t_i$ is mapped to its embedding vector $\mathbf{q}_i$:
\begin{align}
\mathbf{Q} = [\mathbf{q}_1, \mathbf{q}_2, \ldots, \mathbf{q}_{n_{q}}]^\top = f_{\text{norm}}(f_{\text{text}}(q'))  \in \mathbb{R}^{n_{q} \times d_v},
\end{align}
where $f_{\text{text}}(\cdot)$ denotes the text embedding function, including tokenization and embedding lookup, $n_{q}$ is the number of query tokens, and $d_v$ denotes the hidden dimension of the visual encoder. 

This linguistically motivated filtering removes visually uninformative function words 
while preserving semantically meaningful content tokens. Since only a small set of 
high-frequency function words are discarded, the semantic structure of the query 
remains largely intact. The filtering rule is lightweight and can be adapted to 
other languages by adjusting the set of function-word categories.

\noindent \textbf{Text-aware Visual Embedding.}
Given a document image $d$ of size $H \times W$, we feed it into a visual encoder built on a Vision Transformer (ViT)~\cite{dosovitskiy2021imageworth16x16words} backbone. The image is partitioned into non-overlapping patches of size $p \times p$, producing $n_d = \frac{HW}{p^2}$ visual patch tokens. Each patch is processed through the ViT layers, followed by a lightweight projection layer that maps features into a shared embedding space aligned with textual representations, and then normalized for similarity calculation:
\begin{equation}
    \label{doc_embedding}
    \mathbf{D} = [\mathbf{d}_1, \ldots, \mathbf{d}_{n_d}]^\top= f_{\text{norm}}(f_{\text{proj}}(f_{\text{ViT}}(d))) \in \mathbb{R}^{n_d \times d_v},
\end{equation}
where $\mathbf{D}$ denotes the output passage embedding, and $[\mathbf{d}_1, \ldots, \mathbf{d}_{n_d}]^\top$ contains the text-aware visual embedding for each patch.

To bootstrap our visual encoder with strong multi-modal priors, we initialize it from a pretrained multi-modal large language model. 
And to enable the encoder to extract text-rich representations directly from raw document images, we pretrain it following a visual-text alignment objective~\cite{guan2025tokenleveltextimagefoundation}, which leverages token-mask pairs to establish fine-grained correspondence between image patches and textual embeddings.  The encoder is subsequently fine-tuned for document retrieval while preserving learned visual-textual alignment.

\noindent \textbf{Scale-adaptive Late Interaction.} Given query embeddings $\mathbf{Q} \in \mathbb{R}^{n_{q} \times d_v}$ and document embeddings $\mathbf{D} \in \mathbb{R}^{n_d \times d_v}$, we compute a document-level similarity score via late interaction. A standard approach~\cite{khattab2020colbertefficienteffectivepassage} applies max-pooling over document tokens for each query token. However, this is fragile: a single visually similar but semantically irrelevant patch can dominate the score, leading to spurious matches.

We address this with a scale-adaptive aggregation mechanism. For each query token $\mathbf{q}_i$, we select the top-$k$ most similar document patches and average their similarities: 
\begin{equation} 
\mathcal{S}(q, d) = \sum_{i=1}^{n_{q}} \frac{1}{k} \sum\{topK (\{\langle \mathbf{q}_i, \mathbf{d}_1 \rangle ,   \ldots,\langle \mathbf{q}_i, \mathbf{d}_{n_d} \rangle\},k)\}, 
\end{equation} 
where $topK(\mathbf{x}, k)$ is a function
that returns top-$k$ elements in $\mathbf{x}$, and 
\begin{equation} 
k = \max\left(1, \lfloor \gamma \cdot n_d \rfloor\right), \quad \gamma \in (0, 1). 
\end{equation}

This design serves two main purposes. (1) It stabilizes the matching score by aggregating the top-$k$ similarities rather than relying on a single maximum response. This reduces sensitivity to noisy activations or accidental local matches, mitigating score fluctuation and producing a more robust matching signal. (2) It allows $k$ to scale with document complexity. For high-resolution or visually dense documents, a larger $k$ aggregates evidence from multiple relevant regions and further suppresses noise. For simpler documents with fewer patches, a smaller $k$ preserves fine-grained precision and avoids over-smoothing discriminative signals.

\noindent \textbf{Optimization Objective.}
To train the visual encoder to produce embeddings that maximize $\mathcal{S}(q, d)$ for relevant query-document pairs, we employ in-batch contrastive learning. Given a training batch of $B$ query–document pairs $\{(q_i, d_i^+)\}_{i=1}^{B}$, where $d_i^+$ denotes the ground-truth relevant document for query $q_i$, we adopt in-batch contrastive learning. All other documents in the batch serve as negatives for each query. The loss function is formulated as:
\begin{equation}
\label{eq:stage1_loss}
    \mathcal{L}_{sel} = \frac{1}{B} \sum_{i=1}^{B} \log\left(1 +  \exp\left(\text{max}_{j\neq i}\mathcal{S}(q_i, d_j) - \mathcal{S}(q_i, d_i^+)\right)\right),
\end{equation}
This formulation encourages the model to rank the positive document above all in-batch negatives. During training, we apply Low Rank Adaption (LoRA)~\cite{hu2021loralowrankadaptationlarge} to the final ViT layer and the projector, keeping all other parameters frozen. This minimizes training cost while achieving strong retrieval performance.

\subsection{Vision-adaptive Semantic Refinement }
\label{sec:stage2}
While the LLM-free Visual Selection module efficiently narrows the search space from $N$ pages to a manageable candidate set $\mathcal{D}_c$ ($|\mathcal{D}_c| \ll N$), we further introduce MLLM-based refinement to model the semantic nuance. This stage builds upon our proposed visual encoder to inject richer semantic representations into the filtered candidates, circumventing the computational bottleneck of processing the entire corpus. Specifically, we encode the query $q$ with an embedding layer and an LLM. For document $d$, visual features are first extracted via a visual extraction module and adapted via an MLP before being fed into the LLM. Both modalities are ultimately projected into a shared normalized embedding space for similarity computation:
\begin{align}
\label{query_encoding}
\mathbf{Q} &= [\mathbf{q}_1, \ldots, \mathbf{q}_{n_q}] = \mathcal{F}_{\mathrm{query}}(q) \quad \in \mathbb{R}^{n_q \times d_p}, \\
\label{doc_encoding}
\mathbf{D} &= [\mathbf{d}_1, \ldots, \mathbf{d}_{n_d}] = \mathcal{F}_{\mathrm{doc}}(d) \quad \in \mathbb{R}^{n_d \times d_p},
\end{align}
where $d_p$ represents the dimension of the projected vector space. 
The encoding pipelines are defined as $\mathcal{F}_{\mathrm{query}} = f_\mathrm{norm} \circ f_{\mathrm{proj}} \circ f_{\mathrm{LLM}}\circ f_{\mathrm{text}}$ and $\mathcal{F}_{\mathrm{doc}} = f_\mathrm{norm} \circ f_{\mathrm{proj}} \circ f_{\mathrm{LLM}} \circ f_{\mathrm{MLP}} \circ f_{\mathrm{fuse}}$.


Then, we sum the maximum similarity of each query token across all document tokens to compute the similarity score via $\mathcal{S}(q, d)=\sum_{i=1}^{n_q} \max_{1 \leq j \leq n_d} \langle \mathbf{q}_i, \mathbf{d}_j \rangle$.


To prevent excessive parameter growth and redundant computation, the refinement module reuses the visual encoder from the Visual Selection module mentioned in Sec. \ref{sec:stage1} via weight sharing. Furthermore, we introduce an adaptive feature fusion mechanism ($f_{\mathrm{fuse}}$) to integrate visual representations from both stages, fostering effective cross-stage synergy. The detailed design is as follows.

\noindent \textbf{Adaptive Visual Feature Fusion.}
Our architecture shares the lower $1$ to $\ell-1$ ViT layers between the two modules. Given a document image $d$, we first extract the shared intermediate visual features $\mathbf{H}_{<\ell} = \mathcal{T}_{1:\ell-1}(d)$.
For the final $\ell$-th layer, we introduce a dual-branch design. Alongside the text-aware layer $\mathcal{T}_{\ell}^{\text{text}}$ inherited from the Visual Selection module, we instantiate a parallel ViT layer $\mathcal{T}_{\ell}^{\text{layout}}$ initialized from the original pretrained MLLM weights (see Figure~\ref{fig:framework}). This produces two complementary visual feature streams:

\textit{1) Text-Aware Features} ($\mathbf{F}_{\text{text}}$): Inherited from the LLM-free Visual Selection module, these features are optimized for visual-textual alignment and encode rich textual semantics:
\begin{equation}
\label{eq:f_text}
f_{\mathrm{ViT}}: \mathbf{F}_{\text{text}} = \mathcal{T}_{\ell}^{\text{text}}(\mathbf{H}_{<\ell}) \in \mathbb{R}^{n_d \times d_v}.
\end{equation}

\textit{2) Layout-Aware Features} ($\mathbf{F}_{\text{layout}}$): Retaining the general-purpose visual modeling capacity of the original MLLM, these features better capture non-textual structural semantics such as layouts, tables, figures, and whitespace:
\begin{equation}
\label{eq:f_layout}
\mathbf{F}_{\text{layout}} = \mathcal{T}_{\ell}^{\text{layout}}(\mathbf{H}_{<\ell}) \in \mathbb{R}^{n_d \times d_v}.
\end{equation}

Since the text-aware layer is optimized to match textual content, $\mathbf{F}_{\text{text}}$ is highly responsive to text-dense regions. For accurate retrieval, both textual content and document structure (\eg, table layouts, figure positions) are also critical. To reconcile this, we propose an adaptive fusion mechanism that spatially multiplexes the features based on regional characteristics. 

The key insight is that patches in $\mathbf{F}_{\text{text}}$ with low textual alignment naturally exhibit high similarity to  ``space'' semantics. We exploit this property to separate text-dominant regions from background-dominant regions. Specifically, we compute the similarity between $\mathbf{F}_{\text{text}}$ and a predefined whitespace token embedding $\mathbf{q}_{\text{space}} \in \mathbb{R}^{d_v}$ (obtained from the MLLM's vocabulary):
\begin{equation}
\label{eq:whitespace_sim}
\mathbf{s} = \mathbf{F}_{\text{text}} \mathbf{q}_{\text{space}}^\top \in \mathbb{R}^{n_d}.
\end{equation}
We normalize the similarity scores to $[0,1]$ via min-max scaling to yield $\hat{\mathbf{s}}$, and apply an empirical threshold $\tau \in [0,1]$ to generate a binary spatial mask $\mathbf{m}$:
\begin{equation}
\label{eq:region_mask}
\mathbf{m} = \mathbb{I} \left[ \hat{\mathbf{s}} > \tau \right] \in \{0,1\}^{n_d},
\end{equation}
where $\mathbb{I}[\cdot]$ is the indicator function. Here, $m_j = 1$ indicates a background-dominant patch, while $m_j = 0$ corresponds to a text-dominant patch.

Finally, the fused feature $\mathbf{F}_{\text{fused}}$, which is also the output embedding representation of function $f_\text{fuse}$, is computed as:
\begin{equation}
\label{eq:fusion}
f_{\mathrm{fuse}}: \mathbf{F}_{\text{fused}} = (\mathbf{1} - \mathbf{m}) \odot \mathbf{F}_{\text{text}} + \mathbf{m} \odot \mathbf{F}_{\text{layout}},
\end{equation}
where $\odot$ denotes element-wise multiplication with broadcasting along the feature dimension, and $\mathbf{1}$ is a vector of ones. As defined in Eq.~\eqref{doc_encoding}, $\mathbf{F}_{\text{fused}}$ is subsequently fed into the MLP connector and the LLM backbone to output the contextualized document embeddings $\mathbf{D}=f_\mathrm{norm} \circ f_{\mathrm{proj}} \circ f_{\mathrm{LLM}} \circ f_{\mathrm{MLP}}(\mathbf{F}_{\text{fused}})$ for the final similarity computation.

This adaptive fusion strategy provides three distinct advantages: (1) it reuses representations via shared lower layers, avoiding redundant computational overhead; (2) it preserves the sharp text-sensitivity of the Visual Selection module for content-rich regions; and (3) it enriches the overall representation with essential structural and layout clues, yielding a comprehensive document understanding for accurate retrieval.


\noindent \textbf{Hardness-aware Optimization Objective.}
After LLM-free Visual Selection, the candidate set $\mathcal{D}_c$ is enriched with plausibly relevant documents, making the retrieval task significantly more challenging than the initial filtering. Many candidates share similar textual content or visual layouts, requiring the retrieval to learn fine-grained distinctions. Standard contrastive learning treats all negatives equally, which is suboptimal when some negatives are far more confusable than others.

To train MLLM-based Vision-adaptive Refinement to focus on resolving these challenging cases, we adopt a hardness-weighted variant of InfoNCE loss\cite{oord2019representationlearningcontrastivepredictive}. Given a batch of $B$ query–document pairs $\{(q_i, d_i^+)\}_{i=1}^{B}$, we compute all pairwise similarities within the batch. The loss is formulated as:
\begin{equation}
\label{eq:stage2_loss}
    \mathcal{L}_{ref} = -\frac{1}{B} \sum_{i=1}^{B} \log \frac{\exp(\mathcal{S}(q_i, d_i^+) )}{\exp(\mathcal{S}(q_i, d_i^+) ) + \sum_{j \neq i} w_{ij} \cdot \exp(\mathcal{S}(q_i, d_j) )},
\end{equation}
where the hardness weight $w_{ij}$ for each negative pair $(q_i, d_j)$, $j \neq i$, is defined as:
\begin{equation}
\label{eq:hardness_weight}
    w_{ij} = \text{sg}(\exp\left(\mathcal{S}(q_i, d_j)\right)),
\end{equation}
where $\text{sg}(\cdot)$ means computing with stop-gradient to prevent gradient flow.

This formulation adaptively emphasizes hard negatives: documents with high similarity to the query (i.e., confusing cases) receive exponentially larger weights, forcing the model to learn more discriminative representations for challenging distinctions. Easy negatives with low similarity contribute minimally to the loss, allowing efficient training focus on the decision boundary.

During refinement training, we apply LoRA to the transformer layers of the language model backbone and the output projection layer, following the same parameter-efficient strategy as LLM-free Visual Selection part. We freeze the shared ViT layers to maintain the visual-textual alignment learned in LLM-free Visual Selection, and only update the newly added parallel ViT branch, the MLP connector, the language model, and the projection layer via LoRA. This ensures stable training while adapting the model for fine-grained retrieval.

\section{Experiments}
\label{sec:exp}

\noindent \textbf{Implementation Details.}
For LLM-free Visual Selection, we adopt InternViT-300M~\cite{gao2024mini} as the visual encoder backbone. 
The model is fine-tuned with a batch size of 32 for 10 epochs on the training dataset~\cite{faysse2025colpaliefficientdocumentretrieval}. The base learning rate is set to $5\times10^{-4}$. We apply LoRA with rank $r=32$ and scaling factor $\alpha=32$. For Vision-adaptive Semantic Refinement, we build upon InternVL3~\cite{zhu2025internvl3exploringadvancedtraining} as the vision-language backbone. The model is trained with a batch size of 80 for 10 epochs on the same training dataset. 
The base learning rate is also set to $5\times10^{-4}$. LoRA is applied with the same rank configuration. All training experiments are conducted on 4 NVIDIA A800 80GB GPUs. Additional implementation details and hyperparameter analyses are provided in Appendix.

\noindent \textbf{Datasets.}
We conduct experiments on the ViDoRe benchmark~\cite{faysse2025colpaliefficientdocumentretrieval}, 
a large-scale visual document retrieval dataset consisting of diverse document types. ViDoRe provides documents as pure images, which requires joint modeling of textual content, layout structure, and visual cues.
For scalability and latency analysis, we further construct multiple subsets by sampling from the full ViDoRe dataset, 
with sizes ranging from 500 to 7000 document pages. 
These subsets are used exclusively for measuring end-to-end retrieval latency and are not involved in accuracy evaluation.

\noindent \textbf{Metrics.}
We evaluate retrieval performance using Recall@K and NDCG@K, following prior work on visual document retrieval. 
Recall@K measures candidate coverage by checking whether at least one relevant document appears within the top-K results, while NDCG@K evaluates ranking quality by considering both relevance and position. 
We primarily report Recall@100 for the Visual Selection module, 
as its objective is to preserve as many relevant documents as possible within a manageable candidate pool for subsequent refinement. 
For the final retrieval performance, we report NDCG@5 to focus on ranking quality.

\noindent \textbf{Latency Evaluation.}
To evaluate retrieval efficiency, we measure end-to-end query latency including 
query/document encoding, similarity computation, and candidate refinement during retrieval. 
All methods are evaluated on a single NVIDIA A800 80GB GPU with a batch size of 16 under the same dataset and resolution settings to ensure fair comparison.

\subsection{Main Results}
\label{sec:main_results}

\noindent \textbf{Overall Accuracy.}
Table~\ref{tab:main_results} presents the main retrieval results of LightSTAR across ViDoRe sub-datasets. Among the compared methods, LightSTAR achieves the highest overall average NDCG@5 score of 89.1. Compared with conventional text-based OCR pipelines and vision-language contrastive models, LightSTAR achieves substantial gains across all sub-datasets. Among MLLM-based methods, LightSTAR achieves state-of-the-art performance on 5 out of 8 sub-datasets, outperforming the previous best model ColQwen2.5 with an average NDCG@5 of 88.8.
Crucially, LightSTAR achieves this top-tier performance with only 2B parameters, offering substantially better inference efficiency than existing MLLM-based retrievers.

\begin{table*}[t]
\centering
\caption{Performance comparison of LightSTAR with existing methods on the ViDoRe benchmark. Results are presented using NDCG@5(\%). Latency (s) denotes end-to-end retrieval time measured on a corpus of 5{,}000 document pages as reported in Figure~\ref{fig:scaling}. The best results are in \textbf{bold} and the second best are \underline{underlined}.}
\vspace{-0.5em}
\label{tab:main_results}
\resizebox{\textwidth}{!}{
\begin{tabular}{l|c|cccccccc|c|c}
\toprule
\textbf{Method} & \textbf{\#Params} & \textbf{ArxivQ} & \textbf{DocQ} & \textbf{InfoQ} & \textbf{TATQ} & \textbf{AI} & \textbf{Energy} & \textbf{Gov.} & \textbf{Health.} & \textbf{Avg.($\uparrow$)} & \textbf{Latency($\downarrow$)}\\
\midrule
\multicolumn{11}{l}{\textit{Text-based Methods}} \\
BM25~\cite{robertson1994okapi} & -- & 40.1 & 38.4 & 70.0 & 61.5 & 88.0 & 84.7 & 82.7 & 89.2 & 69.3 & --\\
BGE-M3~\cite{chen2025m3embeddingmultilingualitymultifunctionalitymultigranularity} & 569M & 35.7 & 32.9 & 71.9 & 43.8 & 88.8 & 83.3 & 80.4 & 91.3 & 66.0 & --\\
\midrule
\multicolumn{11}{l}{\textit{Vision-Language Contrastive Models}} \\
CLIP~\cite{radford2021learningtransferablevisualmodels} & 151M & 26.5 & 14.6 & 51.7 & 4.7 & 22.9 & 32.4 & 39.8 & 37.5 & 28.8 & --\\
SigLIP~\cite{zhai2023sigmoid} & 883M & 50.2 & 31.3 & 69.7 & 27.5 & 67.8 & 73.5 & 75.3 & 83.1 & 59.8 & --\\
Nomic-Embed-Vision~\cite{nussbaum2024nomicembedvisionexpanding} & 92M & 17.1 & 10.7 & 30.1 &  2.7 & 12.9 & 10.9 & 11.4 & 15.7 & 13.9 & --\\
\midrule
\multicolumn{11}{l}{\textit{MLLM-based Models}} \\
VLM2Vec ~\cite{jiang2024vlm2vec}& 4B & 42.8 & 26.7 & 66.7 &21.4 & 53.5 & 63.5 & 64.0 & 	
70.7 &  51.2 & --\\
E5-V~\cite{jiang2024e5vuniversalembeddingsmultimodal} & 8B & 48.3 & 34.7 & 69.2 & 29.3 & 78.9 & 78.1 & 82.2 & 82.3 & 62.9 & --\\
VisRAG-Ret~\cite{yu2025visragvisionbasedretrievalaugmentedgeneration} & 3B & 80.6 & 42.7 & 85.0 & 49.3 & 94.4 & 92.0 & 87.7 & 92.7 & 78.0 & 1219.9\\
ColPali~\cite{faysse2025colpaliefficientdocumentretrieval} & 3B & 79.1 & 54.4 & 81.8 & 65.8 & 96.2 & 91.0 & 92.7 & 94.4 & 81.9 &\underline{257.3}\\
Nomic-Embed-Multimodal~\cite{nomicembedmultimodal2025} & 3B & \textbf{87.3} &59.6 & 89.8 & 69.4 & 97.5 & 93.3 & 95.7 & 	
97.9 & 86.3 & --\\
ColQwen2.5 & 2B & \underline{87.2} & \underline{63.3} & \underline{91.3} & \underline{79.5} & \underline{98.6} & \textbf{96.3} & 	
\textbf{96.3} & \underline{98.0} & \underline{88.8} & 466.6 \\
\midrule

\multicolumn{11}{l}{\textit{Ours}} \\
\rowcolor{cyan!10}
\textbf{LightSTAR} & 2B & 86.2 & \textbf{63.5} & \textbf{91.7} & \textbf{81.0} & \textbf{99.5} & \underline{95.4} & \underline{95.8}  & \textbf{99.8} & \textbf{89.1} & \textbf{123.9} \\
\bottomrule
\end{tabular}
}
\vspace{-0.5em}
\end{table*}

\begin{figure}
\centering
\includegraphics[width=1\linewidth]{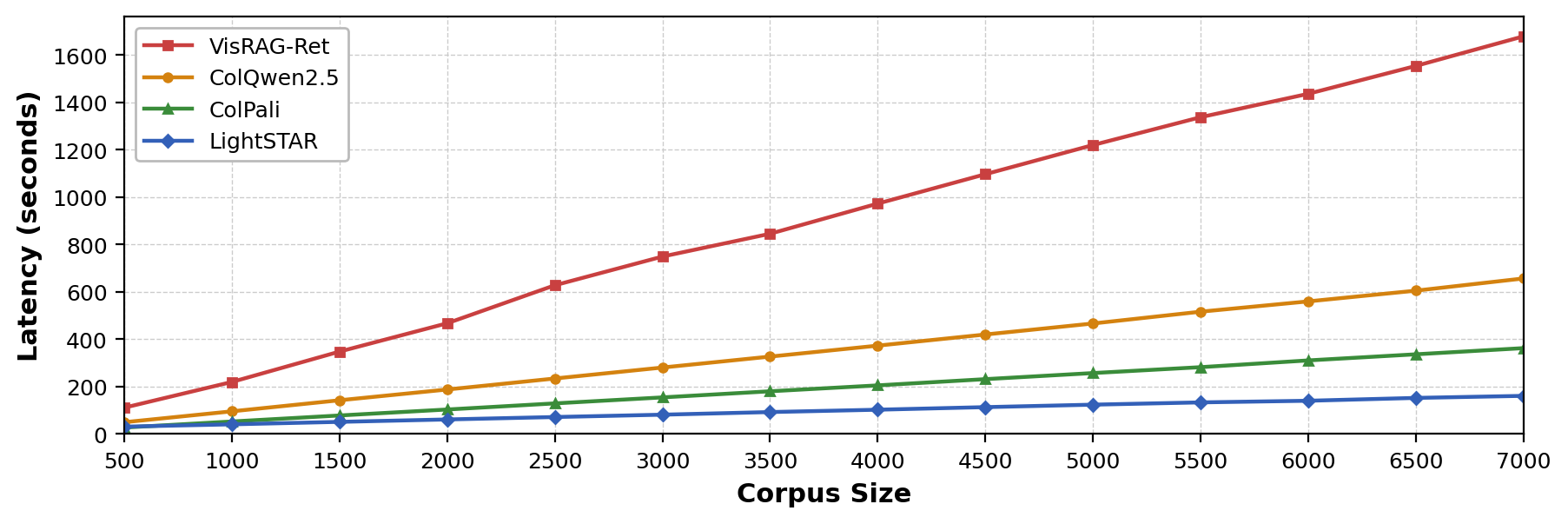}
\vspace{-2em}
\caption{End-to-end retrieval latency comparison between LightSTAR with three competitive MLLM-based retrievers (VisRAG-Ret, ColQwen2.5, and ColPali) as corpus size increases from 500 to 7,000 document pages. 
At 7,000 pages, LightSTAR is 10.4$\times$ faster than VisRAG-Ret, 4.1$\times$ faster than ColQwen2.5, and 2.3$\times$ faster than ColPali, 
demonstrating its computational efficiency for large-scale visual document retrieval.}
\label{fig:scaling}
\vspace{-1.5em}
\end{figure}

\noindent \textbf{Latency and Scalability.}
As shown in Figure~\ref{fig:scaling}, we compare end-to-end retrieval latency and scalability of LightSTAR with three representative MLLM-based methods (VisRAG-Ret, ColQwen2.5, and ColPali), which show competitive retrieval accuracy, to provide a comprehensive scalability analysis. While all methods exhibit an approximately linear increase in latency with corpus size, LightSTAR maintains a much flatter growth than all the baselines, demonstrating superior efficiency and scalability for large corpora.
Its latency increases from 31.1\,s at 500 pages to only 161.1\,s at 7000 pages, yielding the smallest slope among all methods. At 7,000 pages, LightSTAR is 10.4$\times$ faster than VisRAG-Ret (1678.9/161.1), 4.1$\times$ faster than ColQwen2.5 (656.7/161.1), and 2.3$\times$ faster than ColPali (362.6/161.1). These results confirm that our architecture effectively controls computational cost by restricting expensive reasoning to a small top candidate set.


\noindent \textbf{Recall Analysis of the LLM-free Visual Selection Module.}
To highlight the fundamental role of our LLM-free Visual Selection in synergy with the refinement module, we examine its candidate coverage against existing LLM-free methods.
As shown in Table~\ref{tab:recall}, our selection module achieves an average Recall@100 of 97.8, substantially outperforming conventional text-based retrievers and vision-language contrastive models, while remaining fully LLM-free.
These results show that our 542M lightweight visual encoder attains the highest candidate coverage among LLM-free alternatives, preserving nearly all relevant pages within a fixed candidate budget and thus providing an indispensable foundation for subsequent refinement.

\begin{table*}[t]
\centering
\caption{Performance comparison of the LLM-Free Visual Selection module with existing methods on the ViDoRe benchmark. Results are presented using Recall@100(\%) metrics to show the recall coverage. The best results are in \textbf{bold} and the second best are \underline{underlined}.}
\vspace{-0.5em}
\label{tab:recall}
\resizebox{\textwidth}{!}{
\begin{tabular}{l|c|cccccccc|c}
\toprule
\textbf{Method} 
& \textbf{\#Params}
& \textbf{ArxivQ} 
& \textbf{DocQ} 
& \textbf{InfoQ} 
& \textbf{TATQ} 
& \textbf{AI} 
& \textbf{Energy} 
& \textbf{Gov.} 
& \textbf{Health.} 
& \textbf{Avg.($\uparrow$)} \\
\midrule

\multicolumn{11}{l}{\textit{Text-based Methods}} \\
BM25~\cite{robertson1994okapi}   & --    & 71.1 & 70.9 & 90.3 & \underline{96.1} & 98.0 & \underline{99.0} & 95.0 & \textbf{100} & 90.1 \\
BGE-M3~\cite{chen2025m3embeddingmultilingualitymultifunctionalitymultigranularity}  & 569M  & 75.1 & 72.2 & 94.9 & 93.4 & \textbf{100} & 98.0 &  \underline{98.0} & \textbf{100} & \underline{91.5} \\
\midrule

\multicolumn{11}{l}{\textit{Vision-Language Contrastive Models}} \\
CLIP~\cite{radford2021learningtransferablevisualmodels} & 151M & 71.6 & 60.8 & 92.2 & 50.5 & 71.0 & 83.0 & 87.0 & 83.0 & 74.9 \\
SigLIP~\cite{zhai2023sigmoid} & 883M & \underline{89.6} & \underline{72.6} & \underline{95.4} & 84.1 & 89.0 & 96.0 &  \underline{98.0} & \underline{99.0} & 91.2 \\
Nomic-Embed-Vision~\cite{nussbaum2024nomicembedvisionexpanding} & 92M & 66.8 & 53.9 & 71.7 & 41.3 & 49.0 & 52.0 & 46.0 & 50.0 & 53.8 \\
\midrule

\multicolumn{11}{l}{\textit{Ours}} \\
\rowcolor{cyan!10}
\textbf{LLM-Free Visual Selection Module} & 542M & \textbf{96.8} & \textbf{88.9} & \textbf{98.6} & \textbf{98.9} & \underline{99.0} & \textbf{100} & \textbf{100} & \textbf{100} & \textbf{97.8} \\
\bottomrule
\end{tabular}
}
\vspace{-1.5em}
\end{table*}

\subsection{Ablation Studies}
\label{sec:ablation}

\noindent \textbf{Ablation on the Overall Retrieval Architecture.}
Table~\ref{tab:ablation_pipeline} shows a comprehensive ablation results with three configurations: 
(1) selection-only without refinement, 
(2) refinement-only over the full corpus, and 
(3) the complete pipeline. 
The refinement-only setting achieves the highest NDCG@5 of 89.3, 
but incurs a latency increase of nearly three times, since it processes the entire corpus. 
In contrast, the selection-only configuration is significantly faster 
but suffers from a large performance drop (80.8 NDCG@5), 
indicating that visual similarity alone is insufficient for precise ranking. 
In contrast, the full LightSTAR pipeline achieves 89.1 NDCG@5, 
only 0.2 lower than refinement-only, 
while dramatically reducing inference cost. 
This demonstrates that the LLM-free Visual Selection module effectively preserves relevant candidates, 
allowing the refinement module to operate on a compact subset 
with negligible impact on ranking quality and achieving a superior trade-off between effectiveness and efficiency.

\noindent \textbf{Ablation on Module Components.}
Table~\ref{tab:ablation_all} analyzes the impact of different components in different modules.

\noindent \textit{1) Components of the LLM-free Visual Selection Module.}
Removing the content-grounded embedding leads to a 0.8 drop in Recall@100.
This confirms that filtering function words and grounding embeddings in content-bearing tokens effectively reduces spurious visual-text alignments and stabilizes visual matching.
Replacing the proposed scale-adaptive late interaction with a standard late interaction approach~\cite{khattab2020colbertefficienteffectivepassage} which uses max-pooling aggregation causes a 1.0 decrease.
This suggests that simple maximum similarity is highly sensitive to accidental high-scoring patches, resulting in unstable document-level matching.
In contrast, adaptive top-$k$ aggregation mitigates spurious matches and produces more robust similarity estimation.

\noindent \textit{2) Components of the Vision-adaptive Semantic Refinement Module.}
Removing the proposed region-adaptive fusion mechanism without introducing layout-aware features reduces NDCG@5 from 89.1 to 88.3, indicating that the features generated by layout-aware ViT layer serve as complementary signals for text-optimized features. Leveraging the layout/background information captured by layout-aware features with adaptive fusion enables more accurate discrimination between subtle structural differences.
Eliminating the hardness-aware objective reduces performance from 89.1 to 87.9.
This demonstrates that emphasizing hard negatives during training enhances the model’s ability to distinguish highly similar candidate documents.
\begin{table}[t]
\centering
\caption{Ablation study of LightSTAR modules. Performance is measured by average NDCG@5 on ViDoRe and end-to-end retrieval latency(s) measured on a corpus of 5{,}000 document pages. ``Selection'' and ``Refinement'' denote the LLM-free Visual Selection Module and Vision-adaptive Semantic Refinement Module, respectively.}
\vspace{-0.5em}
\label{tab:ablation_pipeline}
\begin{tabular}{cc|c|c}
\toprule
\textbf{Selection} & \textbf{Refinement} & \textbf{ViDoRe} & \textbf{Latency} \\
\midrule
\checkmark & \texttimes & 80.8 & 86.8\\
\texttimes & \checkmark & 89.3 &  465.1\\
\checkmark & \checkmark & 89.1 & 123.9\\
\bottomrule
\end{tabular}
\vspace{-0.5em}
\end{table}

\begin{table}[t]
\centering
\caption{Ablation study of LightSTAR module components. Performance is measured by average Recall@100 and NDCG@5 on ViDoRe.
``CG-Embed'' and ``SA-LateInt'' denote Content-Grounded Embedding and Scale-Adaptive Late Interaction, respectively; 
``Feat-Fusion'' denotes Adaptive Feature Fusion, \texttimes\text{ }means using only text aware features; ``HA-Obj'' denotes Hardness-Aware Objective, \texttimes\text{ }means using Eq.\eqref{eq:stage1_loss} as objective.}
\label{tab:ablation_all}
\vspace{-0.5em}
\begin{minipage}[t]{0.48\columnwidth}
\centering
{\fontsize{7}{8}\selectfont \textbf{(a) Visual Selection Module}}\\[3pt]
\resizebox{\columnwidth}{!}{
\begin{tabular}{cc|c}
\toprule
\textbf{CG-Embed} & \textbf{SA-LateInt} & \textbf{ViDoRe (Recall@100 \%)} \\
\midrule
\checkmark & \texttimes & 96.8 \\
\texttimes & \checkmark & 97.0 \\
\checkmark & \checkmark & 97.8 \\
\bottomrule
\end{tabular}
}
\end{minipage}
\hfill
\begin{minipage}[t]{0.48\columnwidth}
\centering
{\fontsize{7}{8}\selectfont \textbf{(b) Semantic Refinement Module }}\\[3pt]
\resizebox{0.9\columnwidth}{!}{
\begin{tabular}{cc|c}
\toprule
\textbf{Feat-Fusion} & \textbf{HA-Obj} & \textbf{ViDoRe (NDCG@5 \%)} \\
\midrule
\checkmark & \texttimes & 87.9 \\
\texttimes & \checkmark & 88.3 \\
\checkmark & \checkmark & 89.1 \\
\bottomrule
\end{tabular}
}
\end{minipage}
\vspace{-1.5em}
\end{table}

\vspace{-0.5em}
\section{Conclusion}
\vspace{-0.5em}
In this work, we presented LightSTAR, a practical and efficient framework for visual document retrieval that strategically addresses the fundamental accuracy-efficiency trade-off faced by existing MLLM-based approaches. By leveraging the key insight that user queries are typically keyword-anchored, we decompose the retrieval process into two synergistic modules: a lightweight LLM-free Visual Selection module, followed by Vision-adaptive Semantic Refinement. Our extensive experiments demonstrate that LightSTAR achieves state-of-the-art retrieval accuracy while reducing end-to-end latency by several-fold compared to prior MLLM-based methods. Beyond immediate practical impact, LightSTAR establishes a new paradigm for visual document retrieval: rather than applying uniform heavy computation across all documents, intelligent computational allocation based on query characteristics and document properties offers a more sustainable path forward.


\section*{Acknowledgements}
This work was supported by NSFC 62322604 and NSFC 62576207.

%
%
\bibliographystyle{splncs04}
\bibliography{main}

\newpage
\appendix
\title{LightSTAR: Efficient Visual Document \\ Retrieval via Lightweight Selection \\with Vision-Adaptive Refinement \\ (Supplementary Material)} 


\author{Tongkun Guan\inst{1}\textsuperscript{$*$}\orcidlink{0000-0003-3346-8315} \and
Haocheng Wang\inst{1}\textsuperscript{$*$}\orcidlink{0009-0008-8449-1834} \and
Wei Shen\inst{1(\textrm{\Letter})} \and
Xiaokang Yang\inst{1}
}

\authorrunning{T.~Guan et al.}

\institute{MoE Key Lab of Artificial Intelligence, AI Institute, School of Computer Science, \\ Shanghai Jiao Tong University \\
}

\maketitle

\section{Keyword-Anchored Queries Analysis}

To better understand the prevalence of keyword-anchored queries
in document retrieval tasks, we conduct a dataset-level analysis
to examine whether query keywords appear in the visual textual
content of the corresponding ground-truth pages.
This analysis provides empirical support for the assumption used in
Section~\ref{sec:stage1}, namely that user queries often contain
semantically informative tokens that correspond to visual textual
content appearing in relevant document pages.

For each query in the evaluated datasets, we inspect the associated
ground-truth page and determine whether at least one token in the query
appears in the visual textual content of the ground-truth page.
If at least one query token can be identified in the page, the query is
considered to contain visually observable content tokens. This
measurement provides a conservative estimate of whether queries contain
visual textual anchors that can guide document retrieval.


\begin{table}[t]
\centering
\caption{Dataset statistics of keyword-anchored queries, where at least one query keyword appears in the corresponding ground-truth page.}
\label{tab:visual_token_analysis}
\begin{tabular}{lccc}
\toprule
Dataset & Queries & Keyword-anchored Queries & Ratio (\%) \\
\midrule
Energy & 100 & 99 & 99.0 \\
Healthcare Industry & 100 & 100 & 100.0 \\
Artificial Intelligence Test & 100 & 100 & 100.0 \\
Government Reports & 100 & 100 & 100.0 \\
InfoVQA & 494 & 484 & 98.0 \\
TAT-QA & 1646 & 1615 & 98.1 \\
DocVQA & 451 & 423 & 93.8 \\
ArxivQA & 500 & 496 & 99.2 \\
\midrule
Overall & 3391 & 3317 & 97.8 \\
\bottomrule
\end{tabular}
\end{table}

As shown in Table~\ref{tab:visual_token_analysis}, the vast majority of
queries across all datasets contain at least one query keyword appears in the corresponding ground-truth page. In most cases, the required information
appears in visually recognizable elements such as tables, figures, or
structured textual regions. This observation suggests that visual
textual signals provide strong cues for identifying relevant document
pages, supporting the design of the LLM-free Visual Selection module
which leverages lightweight visual-text representations for efficient
candidate filtering.

\section{Hyperparameter Analysis}
In this section, we analyze the impact of several key hyperparameters used in LightSTAR.
Unless otherwise specified, all experiments follow the same configuration as described
in Sec.~4.

\begin{figure}[t]
    \centering
    \begin{subfigure}[b]{0.295\textwidth}
        \centering
        \includegraphics[width=\textwidth]{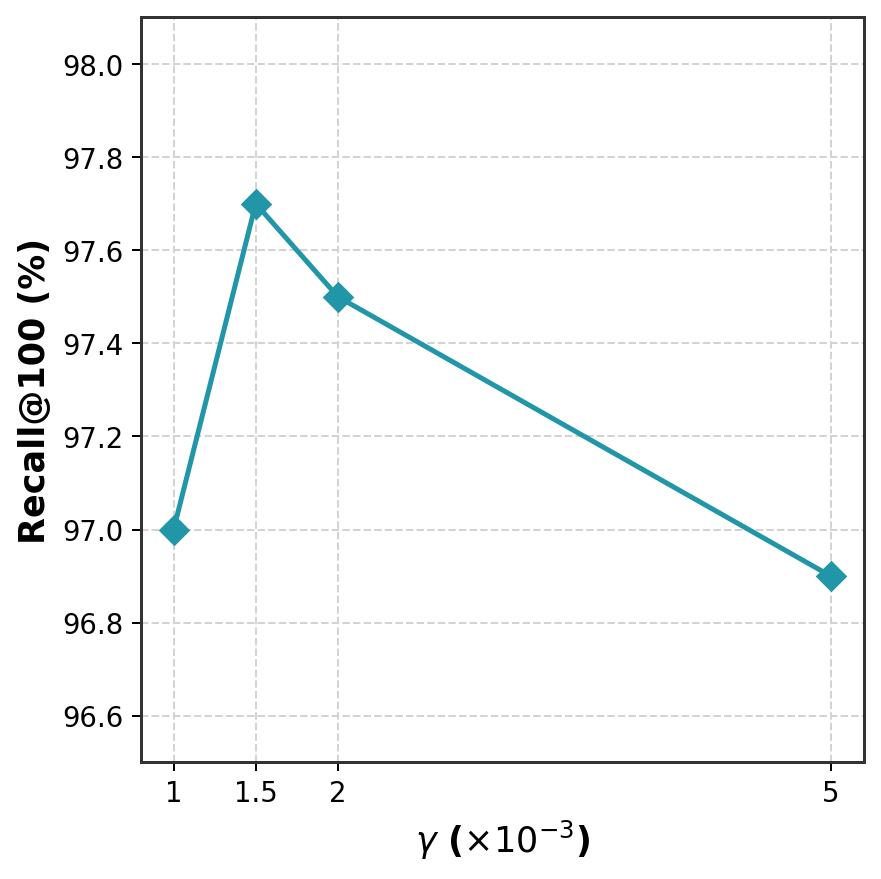}
        \caption{Impact of scale-adaptive parameter $\gamma$.}
        \label{fig:ablation_gamma}
    \end{subfigure}
    \hfill
    \begin{subfigure}[b]{0.33\textwidth}
        \centering
        \includegraphics[width=\textwidth]{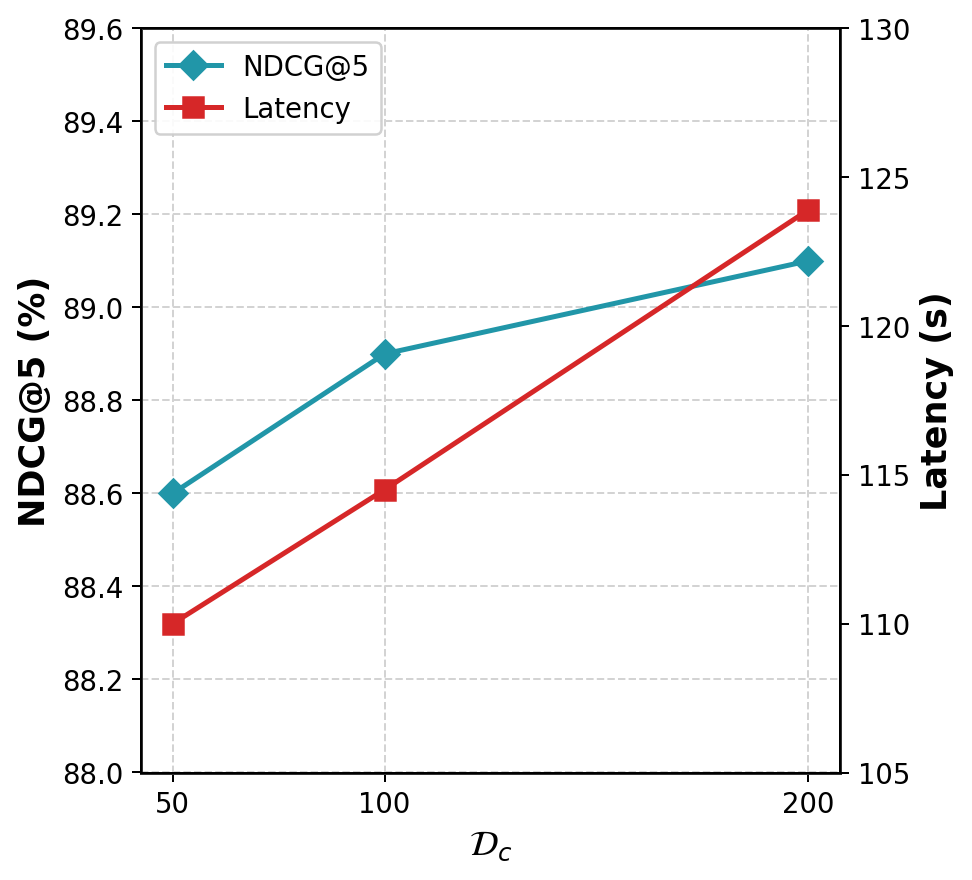}
        \caption{Impact of candidate size $\mathcal{D}_c$ on accuracy and latency.}
        \label{fig:ablation_candidate}
    \end{subfigure}
    \hfill
    \begin{subfigure}[b]{0.298\textwidth}
        \centering
        \includegraphics[width=\textwidth]{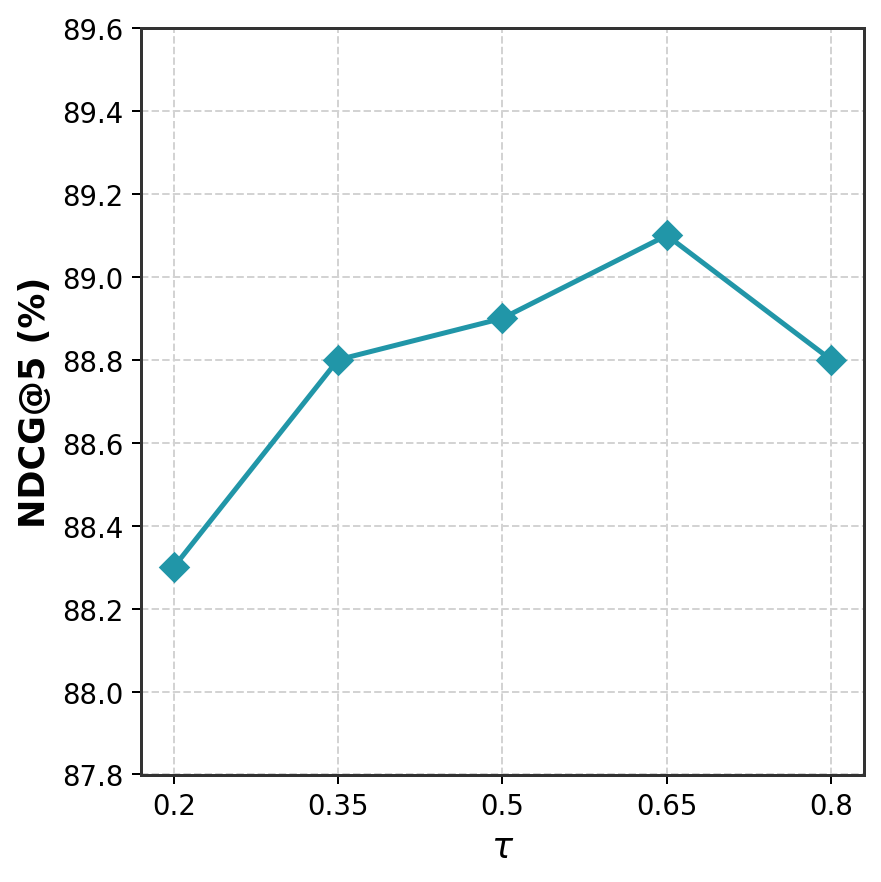}
        \caption{Impact of fusion threshold $\tau$.}
        \label{fig:ablation_tau}
    \end{subfigure}
    \caption{Ablation studies on key hyperparameters.}
    \label{fig:ablation}
\end{figure}

\noindent\textbf{Effect of the Scale-Adaptive Interaction Parameter $\gamma$.}
We first analyze the sensitivity of the scale-adaptive late interaction parameter
$\gamma$ defined in Eq.~(6), which controls the number of document patches
aggregated for each query token.
As shown in Figure~\ref{fig:ablation}\subref{fig:ablation_gamma}, extremely small values of $\gamma$
rely on very few patches and may lead to unstable similarity estimation,
while overly large values introduce excessive smoothing.
LightSTAR achieves stable performance across a reasonable range of $\gamma$.

\noindent\textbf{Effect of Candidate Set Size $\mathcal{D}_c$.}
We further study the influence of the candidate set size $\mathcal{D}_c$ produced by the
LLM-free Visual Selection module.
As shown in Figure~\ref{fig:ablation}\subref{fig:ablation_candidate},
increasing $\mathcal{D}_c$ generally improves retrieval accuracy by allowing the refinement
module to consider more potentially relevant pages. However, larger candidate
sets also increase computational cost. These results demonstrate that the
LLM-free Visual Selection module can effectively reduce the search space
while preserving most relevant candidates for the subsequent refinement stage.

\noindent\textbf{Effect of Fusion Threshold $\tau$.}
We analyze the influence of the fusion threshold $\tau$ in Eq.~(13), which
controls the separation between text-dominant and layout-dominant regions
during adaptive feature fusion.
As shown in Figure~\ref{fig:ablation}\subref{fig:ablation_tau},
the choice of $\tau$ influences the balance between textual semantics and
background layout information. When $\tau$ is set too high, most tokens are
treated as text-dominant regions, which limits the contribution of background
layout cues. In contrast, when $\tau$ is too low, many regions are considered
layout-dominant, leading to a loss of important textual semantics.
A good balance between these two sources of information is achieved when
$\tau = 0.65$.

\section{Comparison with Independent Cascade Baselines}

To further examine whether the effectiveness of LightSTAR simply comes from a
generic multi-stage cascade pipeline, we additionally evaluate independent
cascade baselines that combine an off-the-shelf lightweight visual retriever with
an MLLM-based reranker. Specifically, CLIP and SigLIP are used as the first-stage
retrievers to select candidate pages, and ColQwen2.5 is then applied to rerank
the selected candidates.

As shown in Table~\ref{tab:cascade_baseline}, these independent cascade
baselines are substantially inferior to LightSTAR. Although CLIP+ColQwen2.5 and
SigLIP+ColQwen2.5 follow a similar selection-then-reranking pipeline, their
first-stage retrievers are not specifically optimized for visual document retrieval
or query-patch alignment. Consequently, relevant pages may be missed during
candidate selection, limiting the final reranking performance.

In contrast, LightSTAR adopts a unified selection-refinement design tailored for
visual document retrieval. The LLM-free Visual Selection module and the
Vision-adaptive Semantic Refinement module share the visual backbone, allowing
the visual encoder to support efficient high-recall candidate selection while also
serving as the visual foundation for fine-grained MLLM-based refinement. These
results indicate that the gain of LightSTAR comes not merely from using a
multi-stage cascade, but from the unified architecture and targeted training for
query-patch alignment.

\begin{table}[t]
\centering
\caption{Comparison with independent cascade baselines on ViDoRe. CLIP and
SigLIP are used as independent first-stage retrievers, followed by ColQwen2.5
reranking. LightSTAR achieves better performance with a unified
selection-refinement architecture.}
\label{tab:cascade_baseline}
\begin{tabular}{l c}
\toprule
Method & ViDoRe NDCG@5 (\%) \\
\midrule
CLIP + ColQwen2.5 & 73.3 \\
SigLIP + ColQwen2.5 & 81.3 \\
LightSTAR & \textbf{89.1} \\
\bottomrule
\end{tabular}
\end{table}

\section{Content-Focused Visual-Text Alignment Pretraining}

To better support the content-grounded visual selection paradigm described in the main paper, 
we pretrain the visual encoder with a token-level visual-text alignment objective.
As discussed in Sec.~3, many document retrieval queries are anchored by visually 
observable textual content in the document pages. Therefore, enabling the encoder 
to associate textual tokens with their corresponding visual regions can improve 
the model’s ability to capture query-relevant signals directly from document images.

Following the data construction pipeline introduced in~\cite{guan2025tokenleveltextimagefoundation}, 
we convert document images and their textual transcriptions into token-mask pairs. 
Specifically, the document transcription is tokenized into subword units, and for each 
token we generate a binary spatial mask indicating the image region where the token 
appears. These token-mask pairs provide supervision for learning fine-grained 
correspondence between textual tokens and visual regions.

Different from the original setup in~\cite{guan2025tokenleveltextimagefoundation}, 
we restrict the alignment supervision to \textit{content-focused tokens}. 
Following the query filtering strategy described in Eq.~(2), we first 
identify the content-focused tokens for each query by removing non-informative 
words. During alignment training, we only apply supervision to the tokens that 
match these filtered query tokens in the corresponding ground-truth document page. 
This design encourages the encoder to focus on visually grounded textual signals 
that are directly relevant to retrieval queries.

\begin{figure}
\centering
\includegraphics[width=1\linewidth]{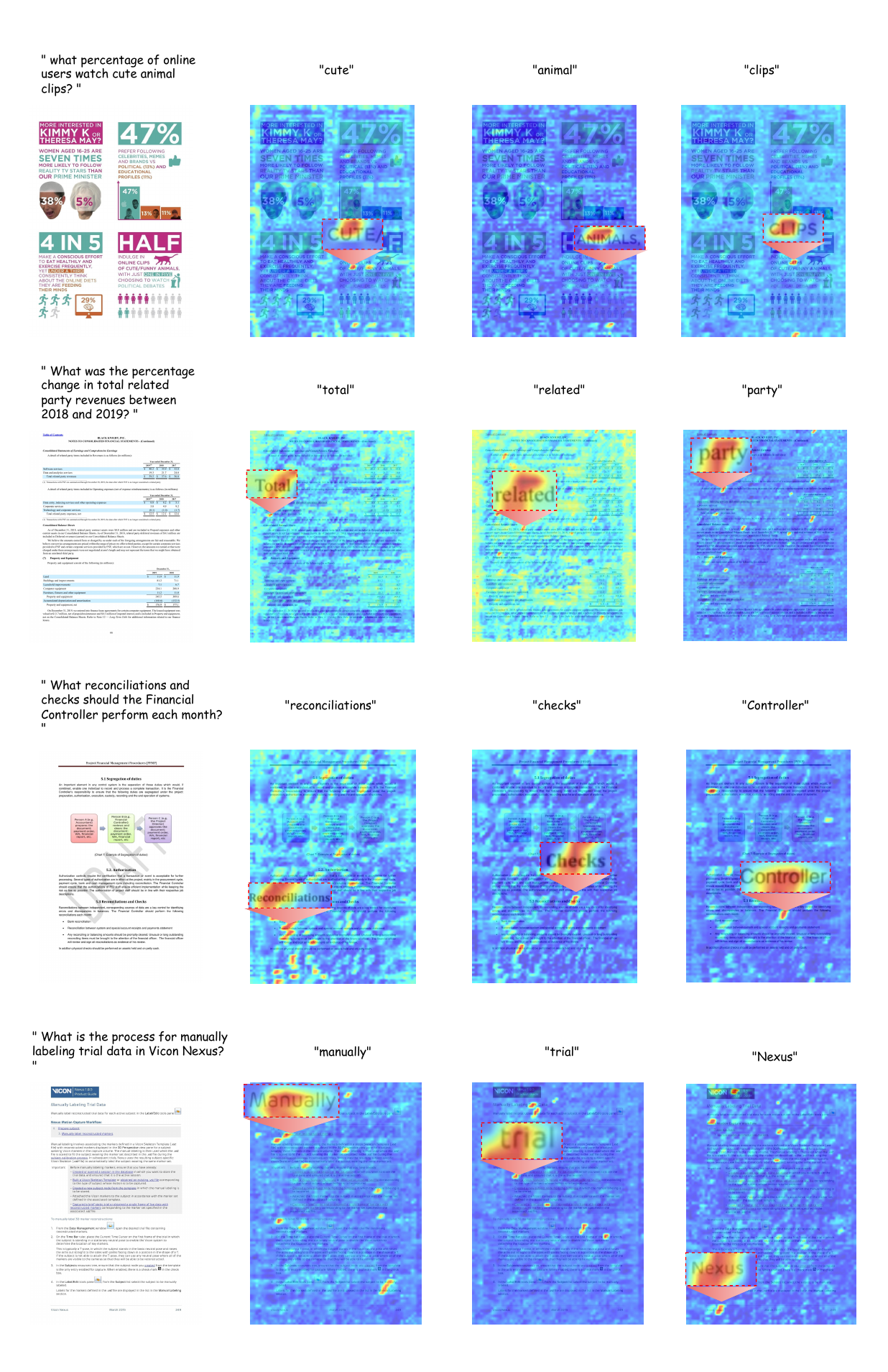}
\vspace{-3em}
\caption{Examples of token-level visual-text alignment.
We visualize the similarity between query token samples and document images.
Warmer colors indicate higher similarity, showing that the visual encoder can 
localize textual regions associated with specific query tokens.}
\label{fig:visualization}
\vspace{-1.5em}
\end{figure}

To qualitatively illustrate the learned visual-text correspondence, we visualize 
the similarity between query tokens and image regions in Figure~\ref{fig:visualization}. 
The highlighted regions indicate areas with high similarity to specific query tokens. 
As shown in the examples, the visual encoder successfully localizes textual regions 
that correspond to the query tokens, demonstrating that the alignment pretraining 
enables the model to capture query-relevant textual signals in document images.

\end{document}